\documentclass[10pt,twocolumn,letterpaper]{article}

\usepackage{cvpr}
\usepackage{times}
\usepackage{epsfig}
\usepackage{graphicx}
\usepackage{amsmath}
\usepackage{amssymb}
\usepackage{arydshln}
\usepackage{subfiles}

\cvprfinalcopy 

\def\cvprPaperID{****} 

\begin{document}

\title{Arithmetic addition of two integers by deep image classification networks: experiments to quantify their autonomous reasoning ability}

\author{Shuaicheng Liu
\and Zehao Zhang
\and Kai Song
\and Bing Zeng\thanks{The corresponding author~(eezeng@uestc.edu.cn)}
\vspace{1.0em}
\and School of Information and Communication Engineering,\\
\and University of Electronic Science and Technology of China\\
}


\maketitle

\begin{abstract}
   The unprecedented performance achieved by deep convolutional neural networks for image classification is linked primarily to their ability of capturing rich structural features at various layers within networks. Here we design a series of experiments, inspired by children's learning of the arithmetic addition of two integers, to showcase that such deep networks can go beyond the structural features to learn deeper knowledge. In our experiments, a set of images is constructed, each image containing an arithmetic addition $n+m$ in its central area, and several classification networks are then trained over a subset of images, using the sum as the label. Tests on the excluded images show that, as the image set gets larger, the networks have well learnt the law of arithmetic additions so as to build up their autonomous reasoning ability strongly. For instance, networks trained over a small percentage of images can classify a big majority of the remaining images correctly, and many arithmetic additions involving some integers that have never been seen during the training can also be solved correctly by the trained networks.
\end{abstract}

\section{Introduction}
Human intelligence is mostly represented by the ability of drawing inferences on many other cases (similar or new) from the learnt instances or knowledge~\cite{gottfredson1997mainstream,van2009efficiency}. Such ability can be built up gradually for nearly every human being during his/her learn-and-growth process. One of the most typical scenarios is the learning of the arithmetic addition of two integers $n+m$ with children at very early ages. During the learning, parents/teachers show examples one by one and tell the answers and then test children with new additions. One proven result~\cite{groen1977can} is that, after teaching several dual-examples such as ``$1+2$, $2+1$", ``$1+3$, $3+1$", and ``$2+3$, $3+2$", and additionally $1+4$, some children could give the correct answer to the new addition $4+1$. At this stage, children may not understand the commutativity of arithmetic additions at all. Nevertheless, they get the correct answer merely by a simple inferencing. As the learning continues, children can understand the law of arithmetic additions gradually and then conduct new additions correctly, which builds up their autonomous calculating ability~\cite{gilmore2007symbolic}.

\begin{figure*}[t]
  \centering
  \includegraphics[width=1.0\linewidth]{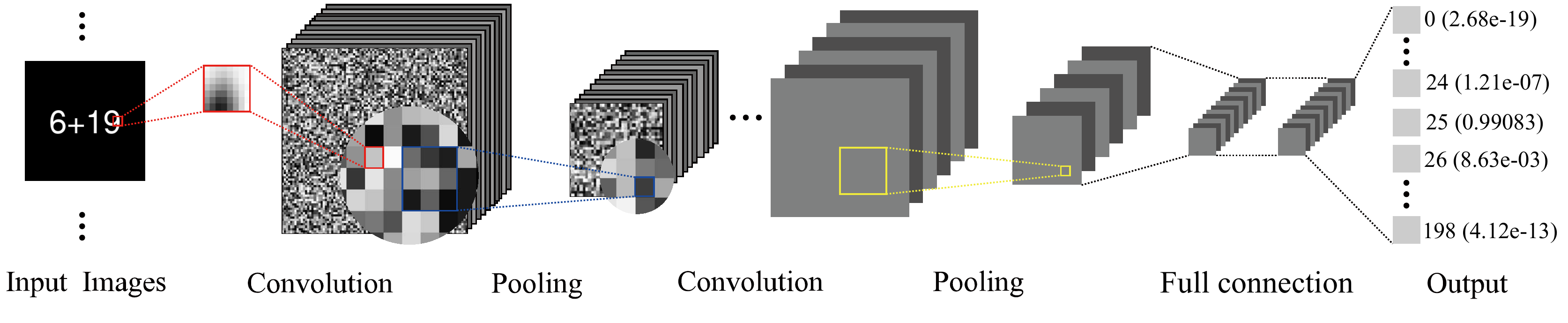}
  \caption{\textbf{The CNN architecture for the arithmetic addition of two integers}. The network sees the input images only, whereas no knowledge (such as the arithmetic interpretation of an integer, the meaning of `+', or the rules of arithmetic additions) is provided. The output placed at the right-end represents the probabilities of belonging to various classes (labeled by $0\sim198$), which is obtained from a real test ($6+19$) - the highest probability ($0.99989$) occurs at $25$ (corresponding to the correct result).}\label{fig:intro}
\end{figure*}

Here, each example is usually presented on a white board or a piece of paper. Children observe the question and start the learning. They first recognize two integers $n$ and $m$ as well as the symbol `+', through their structural patterns. Then, they will learn the law of arithmetic additions gradually when more and more examples are taught. Clearly, such law is far beyond and thus much more important than the structural patterns perceived from the arithmetic addition formula itself, as it implies deeper and more intrinsic knowledge.

In comparison, we have witnessed over the past decade a huge success of various deep learning networks (DLNs) in applications such as image analysis, machine translation, and natural language processing~\cite{lecun2015deep}. One benchmark example is the worldwide image-classification competition around ImageNet~\cite{krizhevsky2012imagenet,deng2009imagenet}, where deep convolutional neural networks (CNNs) have been dominant for years. The unprecedented performance achieved by these classification networks is linked primarily to their ability of capturing rich structural features at various layers within networks~\cite{zhang2016understanding}. Nevertheless, as we are aware so far, no examples have ever been given to showcase whether or not and then to what extent these deep networks have learnt intrinsic knowledge that is beyond the structural features perceived from individual images~\cite{keskar2016large}. This is largely because it is extremely challenging to define clearly the specific knowledge that needs to be learnt in the image classification task (as well as many other similar tasks), which thus hinds the quantitative assessment on the network¡¯s autonomous reasoning ability after the training is completed.

Encouragingly, the recent work around AlphaGo~\cite{silver2016mastering,silver2017mastering,singh2017artificial} and AlphaStar~\cite{vinyals2019grandmaster} has shone light on providing some proven evidence to the reasoning ability of DLNs. By supervised learning and reinforcement learning, these Alpha-agents have almost swept top-ranked human players. However, these successful examples require that the game-rules be provided explicitly.

\begin{figure*}[t]
  \centering
  \includegraphics[width=1.0\linewidth]{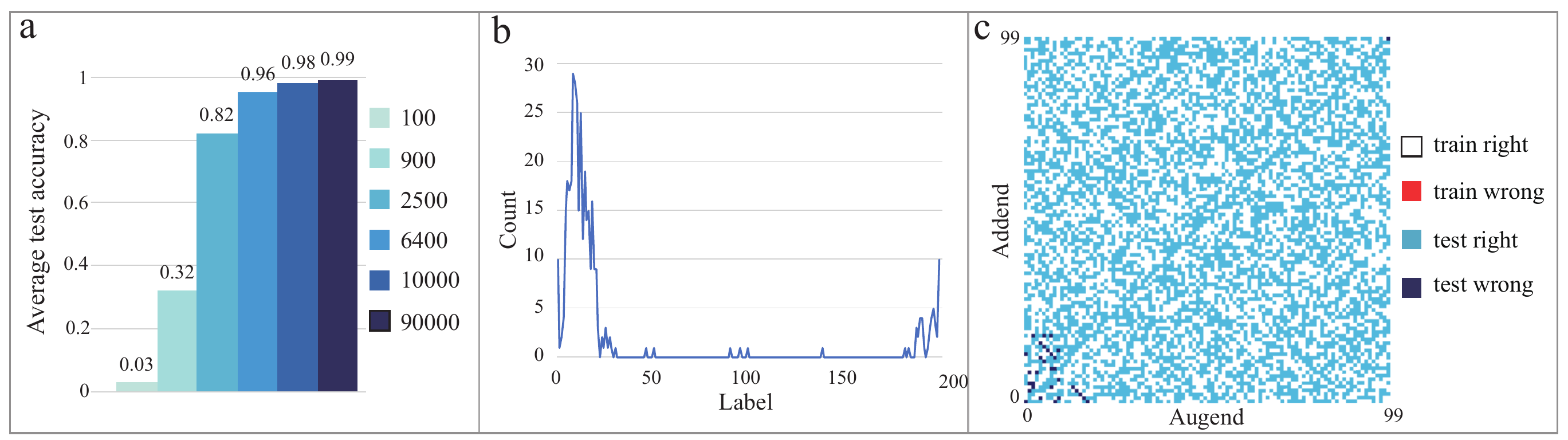}
  \caption{\textbf{Verification of the commutativity law of arithmetic additions}. While all images $n+n$ are put into $\Psi$, the image set $\Omega$ is split in such a way that whenever an image $n+m (n\neq m)$ is selected into $\Phi$, its dual image $m+n$ is surely put into $\Psi$, leading to the $50\%-50\%$ split. For each image set, the experiment is conducted independently for $10$ times. \textbf{a.} Test accuracy averaged over 10 trials. The image set needs to be large enough so that the commutativity law can be well learnt (with an accuracy at $90\%$ or above). \textbf{b.} Test error distribution of the image set $|\Omega|=10,000$ $(n,m \leqslant99)$. We count the error times and accumulate them over $10$ trials. Errors are highly clustered around two ends (with small or big labels). \textbf{c.} Learning map of one trial over the image set $|\Omega|=10,000$. Here, the training is 100\% accurate as no cells of red color appear. Calculations of $n+n$ are all correct, except for $0+0$ and $99+99$. All incorrect inferences (cells of deep-blue color) occur around the corner ($0,0$).}\label{fig:commutative}
\end{figure*}

\section{Autonomous reasoning and beyond}
We return to deep image classification networks where no rules or knowledge are provided, except for the labelling of all training data. Here, we focus on two questions: (1) how strongly can these networks build up their autonomous reasoning ability after the training is completed and (2) have they learnt specific knowledge that is beyond the structural features so as to lead to a strong autonomous reasoning ability. Inspired by children's learning of the arithmetic addition of two integers, as discussed earlier, we carry out a series of experiments on several deep CNNs for image classification to assess their autonomous reasoning abilities quantitatively.

To this end, we construct a closed-set of images, each image containing an arithmetic addition of two integers $n+m$ in its central area, see the input part of Figure~\ref{fig:intro} for one example. Then, we train several deep CNNs over a subset of images. Although each image in our experiments is very simple as compared to the images used in ImageNet, it does imply clear rules or knowledge that are easy to understand for everyone when observing the image. More specifically, the goal here is threefold: (1) to recognize two integers and the symbol `+' within each formula, (2) to understand the arithmetic meaning of each integer and the symbol `+' (more important), and (3) to master the law of arithmetic additions (most important): ones digits versus tens digits, commutativity between addend and augend, and carry-over in additions. The first part is a classic pattern recognition task. The second and third parts together form the learning of necessary knowledge for computing arithmetic additions.

After the training is completed, we test all images that are not seen during the training. Clearly, the test results will illustrate whether or not and then to what extent these networks can learn the specific knowledge (for mastering arithmetic additions) that is beyond the structural patterns perceived from images. Only under the circumstance where the necessary knowledge has well been mastered can strong autonomous reasoning abilities be built up for the trained networks.

We would like to point out that some of the previous works have addressed the task of arithmetic additions, but in different ways reflecting different properties of networks. For example, Liang \etal applied Optical Character Recognition (OCR) system to recognize operators as well as the digits from images, based on which the additions are calculated accordingly~\cite{liang2016character}. Hoshen \etal adopted a network that consists of only fully connected layers to conduct the task of additions~\cite{hoshen2016visual}, where inputs to the network are two images and the output is also an image containing the summation result, and a simple OCR system is used for recognizing digits for measuring the accuracy. As a result, the network learns the function of image-to-image mapping that concentrates on the transformation of structural patterns perceived in images. In contrast, we directly output the results as the classification labels rather than output images, allowing the network to focus on the learning of the underlying knowledge of arithmetic additions as well as its autonomous reasoning ability.

\section{Image set and our experiments}
Let $\Omega$ denotes an image set and $|\Omega|$ its size (i.e., the number of images in the set). By upper-limiting $n$ and $m$, we have constructed several image sets whose sizes range from $100$ to $90,000$. Here, each image is of resolution $224 \times 224$.

By using the sum as the label for each image, we can convert the calculation of n+m into an image classification problem. We have trained several popular classification networks, such as VGG~\cite{simonyan2014very}, ResNet~\cite{he2016deep}, and SENet~\cite{hu2018squeeze}, see Figure~\ref{fig:intro} for the general architecture shared by these classification networks. Instead of measuring the Top-5 accuracy in the traditional image classification task, we measure the Top-1 accuracy in our experiments, as we are facing a scientific calculation in which the absolute accuracy is the top-priority requirement.

We split the image set $\Omega$ into the training set $\Phi$ and test set $\Psi$. We first construct $\Phi$ by a random selection. Here, we focus on finding out how many images are needed for the training so that a big majority ($90\%$ or above) of the test images can be classified correctly by the trained network. Then, we construct $\Phi$ by excluding all arithmetic additions $n+m$, where $n$ or $m$ $\in[L, U]$, i.e., any integer in the interval $[L, U]$ will not appear at either addend or augend or both. In this way, we would like to showcase that, although those integers have never been seen by networks during the training, a big majority of arithmetic additions involving them can still be calculated correctly.

For each experiment, we train the network independently for $10$ times. Here, we only report the results of SENet, as very similar results have been obtained in other networks such as VGG and ResNet.

\section{Results on randomly-selected subsets}
In the first experiment, we try to verify the commutativity between addend and augend in arithmetic additions. To this goal, we remove all images that contain $n+n$ (i.e., they are all put into the test set $\Psi$). For other images, we select randomly some $n+m$ images $(n\neq m)$ into the training set $\Phi$ but make sure that their dual $m+n$ images are all excluded. In this way, we can select maximally $50\%$ of images into the training set.

After the training is completed, we first examine the calculation of $n+n$. The results show that we can achieve the $100\%$ accuracy when the image set is large enough ($|\Omega|=900$ or above), with two exceptions only, i.e., $0+0$ and $N+N$ ($N$ denotes the maximum integer), as these two additions do not have the corresponding labels.

Then, we test all dual images. Figure~\ref{fig:commutative}a presents the accuracy averaged over 10 independent trials. When the image set gets large, the accuracy increases dramatically, e.g., to almost $100\%$ when $|\Omega|=90,000$. Consequently, we believe that the commutativity law (the knowledge beyond the structural features) should have been learnt very successfully.

Next, we focus on analyzing incorrect calculations in the image set with $|\Omega|=10,000$ (as an example). To this end, we count the numbers of incorrect inferences and accumulate them over 10 trials, with the results shown in Figure~\ref{fig:commutative}b. As seen, most incorrect calculations are gathered at two ends with small or big sums (labels). This is because that there are fewer images in $\Omega$ at these two ends, thus leading to fewer training images to be observed by the network during the training, so that the reasoning logic cannot be built up robustly. We further verify the above distribution by looking at the results of one trial. Figure~\ref{fig:commutative}c shows the learning map of one trial, where each small cell represents 1 out of 4 possible states (in different colors): train right, train wrong, test right, and test wrong. As seen, no cells of red color appear (i.e., the training is $100\%$ accurate); no cells of deep-blue color appear along the 45$^\circ$ diagonal line (i.e., calculations of $n+n$ are all correct, except for $0+0$ and $99+99$); and only 30 cells of deep-blue color (incorrect inferences) appear, all gathering around the corner $(0,0)$.

\begin{figure}[t]
  \centering
  \includegraphics[width=1.0\linewidth]{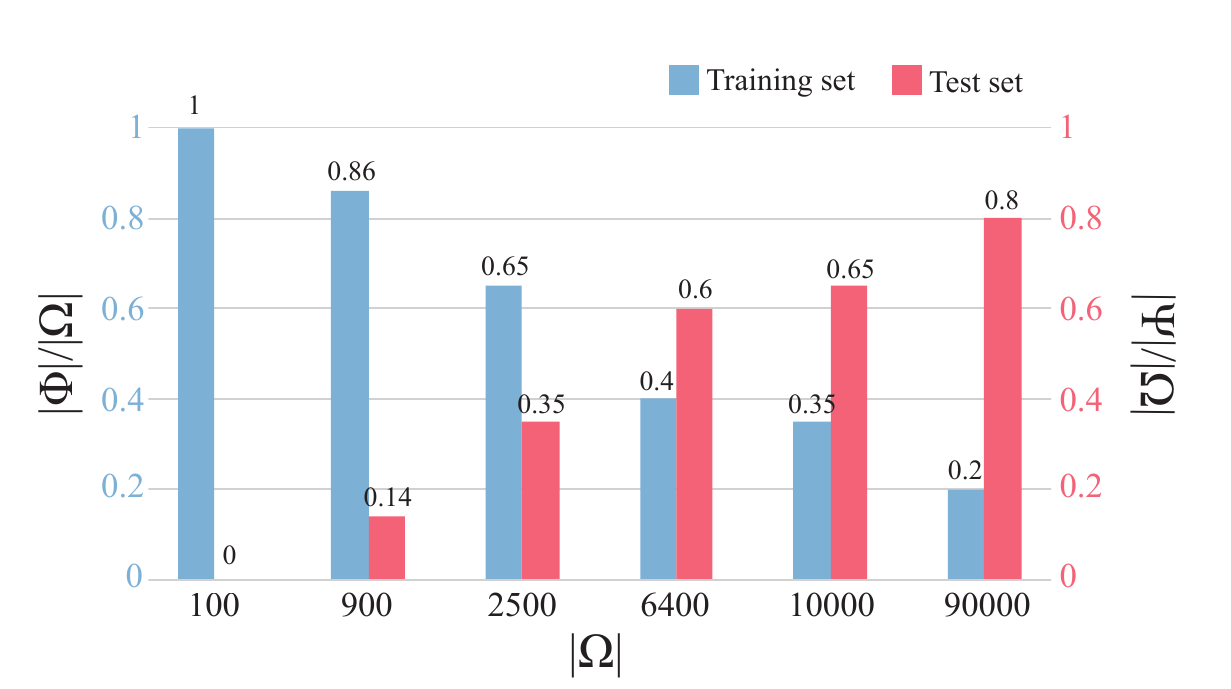}
  \caption{\textbf{Training set versus test set for achieving over $90\%$ test accuracy}. Randomly-selected pairs of $n+m$ and $m+n$ are excluded from the training. Bars of blue-color (training set in percentage with respect to the whole image set) decrease when $|\Omega|$ becomes larger. At the same time, bars of red-color (test set in percentage with respect to the whole image set) increase.}\label{fig:fig3}
\end{figure}

In the second experiment, we also remove all images that contain $n+n$. Then, we further exclude randomly some pairs of $n+m$ and $m+n$ images so that they will never be seen during the training. Here, we aim at finding out how many images are needed for the training so that a big majority (over $90\%$) of the test images can be classified correctly by the trained networks. This is equivalent to how to split the image set $\Omega$ into the training set $\Phi$ and test set $\Psi$, in percentages. Figure~\ref{fig:fig3} shows these percentages for image sets of different sizes so that the test can achieve over $90\%$ accuracy. As seen, when the image set size is too small, i.e., $|\Omega|=100 (n, m\leqslant9)$, we cannot even exclude a single pair from the training as they cannot be calculated correctly. In other words, the learning of arithmetic additions involving one-digit integers fails. When $|\Omega|=900 (n, m \leqslant29)$, the task becomes solvable, whereas $86\%$ images in $\Omega$ needs to be selected into the training set so as to achieve over $90\%$ accuracy on the test set. For the largest image set $|\Omega|=90,000 (n, m \leqslant299)$, only $20\%$ images in $\Omega$ are needed for the training in order to accomplish the same task.

In the third experiment, we select images in $\Omega$ to form the training set $\Phi$ completely randomly (i.e., without any constraints). Here, we focus on the dynamic relationship between the test set size and the test accuracy. Figure 4 shows the results (again, averaged over $10$ trials), where $|\Psi|/|\Omega|$ implies how many images in $\Omega$ are used in tests (in percentage). As seen, for $|\Omega|=100$, the test accuracy drops quickly when more images in $\Omega$ are reserved into the test set $\Psi$. This situation improves when $|\Omega|$ increases. For instance, for $|\Omega|=90,000$, a very high test accuracy (over $90\%$) can be maintained even when we only select $15\%$ of images in $\Omega$ to train the network.

\begin{figure}[t]
  \centering
  \includegraphics[width=1.0\linewidth]{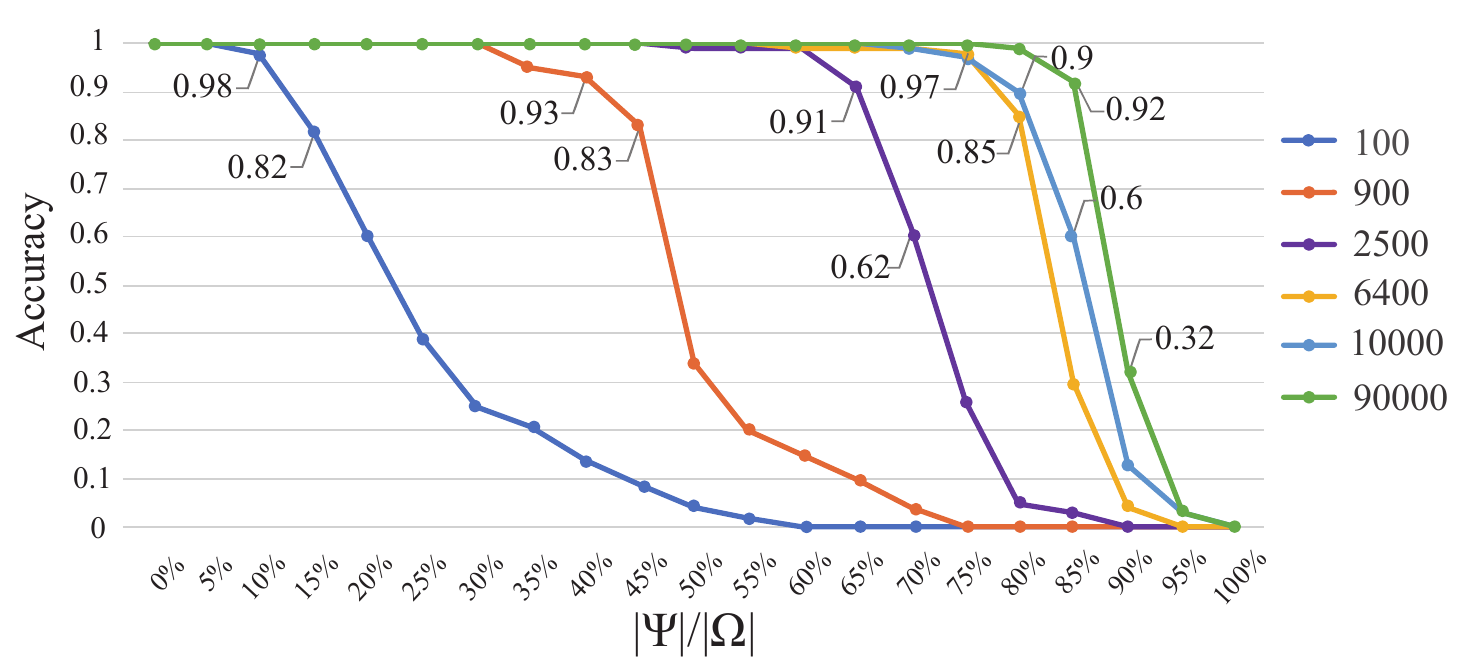}
  \caption{\textbf{Dynamic relationship between the test set size and the test accuracy}. For each image set, the test accuracy drops when more images in $\Omega$ are reserved for test (because fewer images are used for the training). When $|\Omega|$ is small, the curve drops quickly. As $|\Omega|$ gets large, a very high test accuracy (e.g., $90\%$ or above) can be maintained over a big range.}\label{fig:fig4}
\end{figure}

\section{Results on regularly-selected subsets}
In the following experiments, we obey certain rules to exclude some arithmetic additions from the training set. First, we exclude all arithmetic additions of form $n+X$ or $X+n$ from the training. Here, $n$ is a fixed integer and $0\leqslant X\leqslant N$. This task seems very easy as our multiple trials (choosing different $n$) show that the trained network can classify all excluded images correctly. This result indicates that although the network is blind to one specific integer, a
very strong reasoning logic has also been built up after the training is completed.

\begin{figure*}[t]
  \centering
  \includegraphics[width=1.0\linewidth]{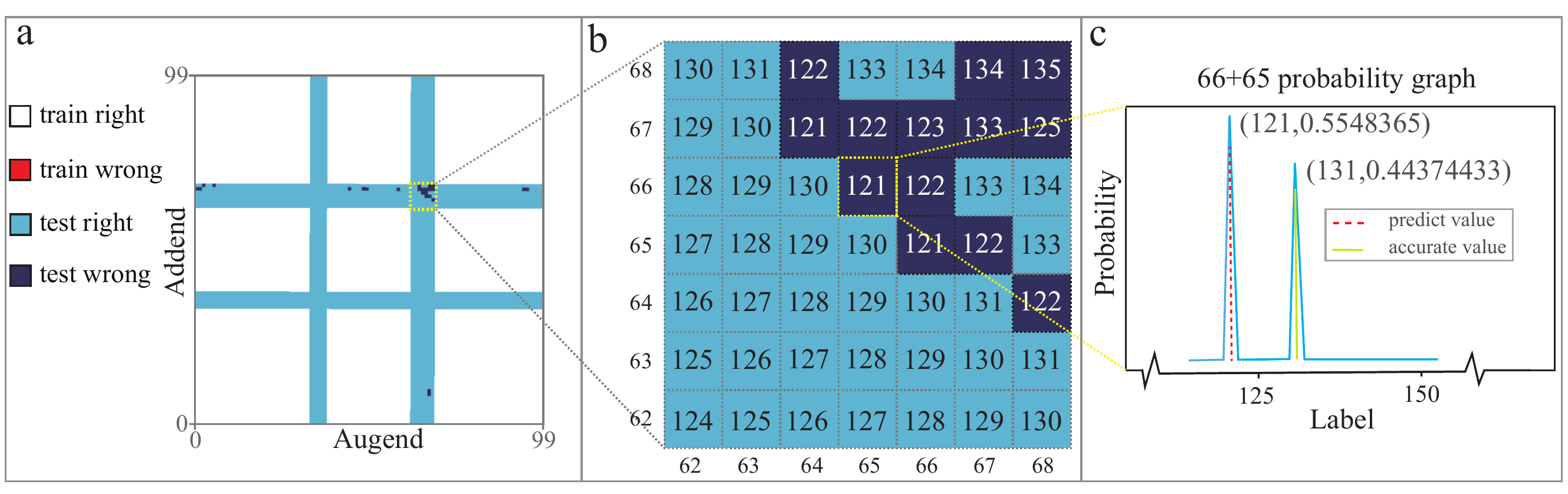}
  \caption{\textbf{Well-trained networks can solve arithmetic additions involving some integers that have never been seen during the training}. Arithmetic additions of form $n+X$ or $X+n$, where $n\in[33, 37]$ and $n\in[62, 68]$ $(0\leqslant X\leqslant 99)$, are all excluded. \textbf{a.} The training is $100\%$ correct (as no cells of red color appear), but a few test errors happen (at cells of deep-blue color). \textbf{b.} One intersection region (with most test errors) is highlighted. \textbf{c.} For the cell $66+65$, the classification produces an incorrect result $121$ with the highest probability $0.5548365$, whereas the correct answer 131 receives a probability $0.44374433$ (the second highest).}\label{fig:fig5}
\end{figure*}

To increase the level of difficulty, we exclude several consecutive integers. For instance, arithmetic additions of form $n+X$ or $X+n$, where $n\in[33, 37]$ and $n\in[62, 68]$ $(0\leqslant X\leqslant 99)$, are all excluded, as shown in Figure~\ref{fig:fig5}. Here, Figure~\ref{fig:fig5}a shows the learning map that displays 4 types of cells, i.e., train right, train wrong, test right, and test wrong (the same as Figure~\ref{fig:commutative}c). Notice that two color strokes in this figure form the test set so that each image in this set has been excluded from the training but needs to be classified during the testing.

In the learning map, an intersection region (highlighted by the yellow rectangle) produces a few incorrect results Then, we zoom-in this region to examine the detailed results, as shown in Figure~\ref{fig:fig5}b. Interestingly, most of those incorrect calculations lost 10, indicating that the network seems not fully understand the carry-over rule in arithmetic additions. We further zoom into one incorrect calculation ($66+65$) to examine the distribution of all classification probabilities, as shown in Figure~\ref{fig:fig5}c. It is found that the network produces $121$ as it receives the highest probability ($0.5548365$), whereas the correct answer 131 receives the second highest probability ($0.44374433$).

Finally, we would like to report that the network will collapse if integers are excluded consecutively in a range such as $n\in[60, 69]$. We believe that this is because the integer 6 has never been seen by the network in the tens position, whereas cases of seeing it in the ones position do not help the learning on the tens position.

\section{Discussion}
Some discussions are necessary as complementary to the quantitative assessment results presented above. First of all, we would like to point out that the labelling of training data itself does contain rules of arithmetic additions. Our results show that these rules have been well learnt so that new arithmetic additions can be solved (with a very high accuracy) by the trained networks.

Our discussions in the following focus on a comparison between the deep CNN-based image classification and the learning of arithmetic additions by image classification networks, with respect to the mechanism in these two tasks. In the former one, structural features extracted from images play the most important role solely. Taking ImageNet as an example, an image is classified into the class ``dog" because neuron that extract features such as dog's legs, dog's head, and etc. are triggered. Similarly, for the class ``bird", neuron taking care of different parts of birds (e.g., eyes, wing, legs, and etc.) are triggered.

\begin{figure}[t]
  \centering
  \includegraphics[width=1.0\linewidth]{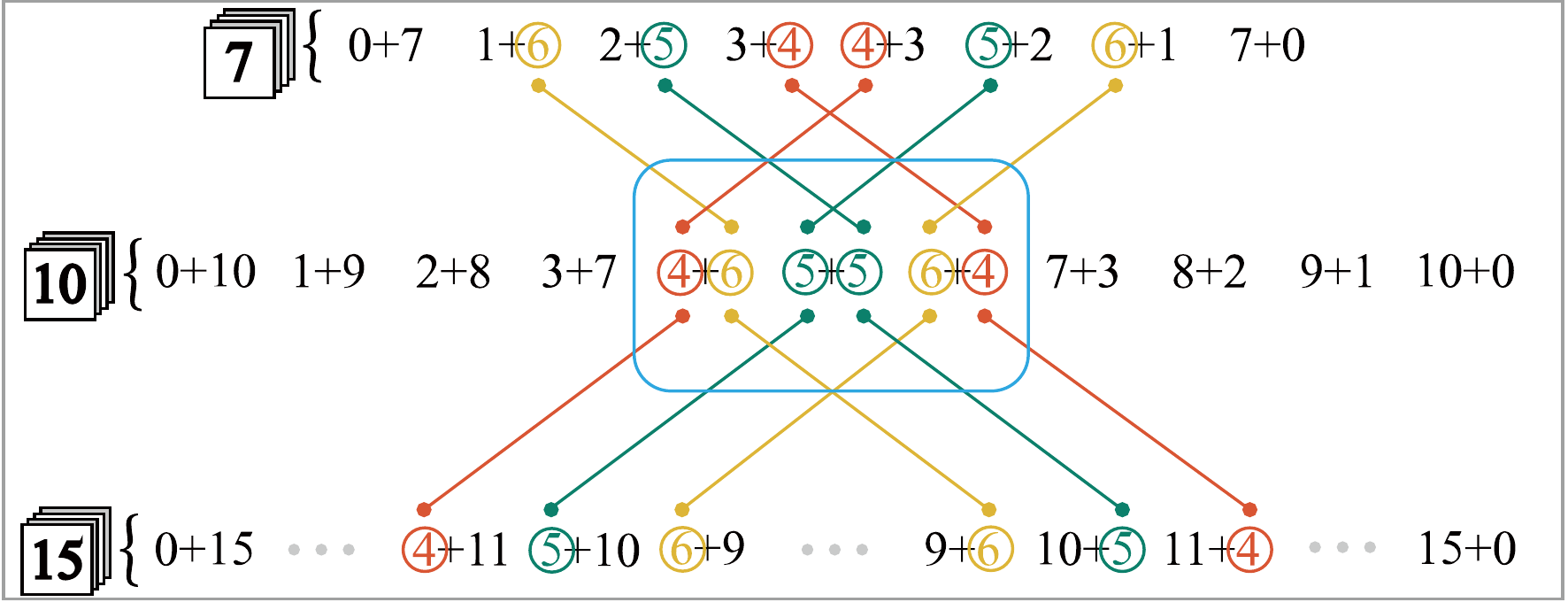}
  \caption{\textbf{Strong autonomous reasoning abilities are linked to successful learning of the task-specific knowledge.} Three combinations of 4, 5, and 6 (enclosed by the blue-color box) are not seen during the training so that the corresponding patterns do not contribute to the learning of Class-10. Nevertheless, the correct test results prove that the trained networks indeed have strong autonomous reasoning abilities, which implies that the law of arithmetic additions should have been well learnt after the training is completed.}\label{fig:fig6}
\end{figure}

In the later task, however, we see a different scenario. As illustrated by the example shown in Figure~\ref{fig:fig6}, 11 different combinations of two integers constitute Class-10. If $5+5$ and $4+6/6+4$ are taken away from the training (which does happen in our experiments presented earlier), the structural patterns of 4, 5, and 6 do not contribute to the training of this class. Seeing one of them (in combination with another integer) during the training would lead to a classification to other classes, e.g., $\textcircled{4}+3=7$ (or $3+\textcircled{4}=7$), $\textcircled{5}+8=13$ (or $8+\textcircled{5}=13)$, and $\textcircled{6}+9=15$ or ($9+\textcircled{6}=15$).  Now, when testing $4+6$ (or $6+4$) and $5+5$, any mechanism driven by structural patterns would more likely classify each of them into a class that saw 4 or 6 or 5 during the training, rather than Class-10 that never sees any of them during the training.
Nevertheless, our experimental results presented earlier show that a big majority of those arithmetic additions excluded from the training has been classified correctly. We believe that this owes to the successful learning of the knowledge (that is beyond the structural patterns perceived from individual images) involved in computing arithmetic additions so that the trained networks have built up their autonomous reasoning ability strongly. To the best of our knowledge, it is the first time that a meaningful example be designed as the proving evidence to the learning of deeper knowledge that can be defined clearly but is far beyond the structural features perceived from individual images.

{\small
\bibliographystyle{unsrt}
\bibliography{deepaddition}
}

\end{document}